\documentclass[twocolumn,twoside,showkeys,preprintnumbers,amsmath,amssymb,secnumarabic, nofootinbib, nobibnotes,epl]{cbpf}
\usepackage[a4paper,dvips=false,pdftex=false,vtex=false]{geometry}
\usepackage{epsfig,epsf}
\usepackage{graphicx}
\usepackage{dcolumn}
\usepackage{bm}
\usepackage{fancyhdr}
\usepackage{pslatex}
\usepackage{pstricks}
\usepackage{graphicx,url}       	  	

\usepackage[english]{babel}  
\usepackage[latin1]{inputenc}   			
\usepackage{setspace}       					
\usepackage{indentfirst}   	 					
\usepackage{enumerate}      					
\usepackage{tikz}                   	
\usepackage{color}
\usepackage{amsmath}
\usepackage{siunitx}
\usepackage{refcount}

\usepackage[nottoc]{tocbibind}
\usepackage{textcomp}
\usepackage{gensymb}

\begin{document}

\language=0
\linespread{1.0}

\noindent
{\footnotesize dx.doi.org/10.7437/NT2236-7640/2017.01.003} \newline
{\footnotesize Notas T\'ecnicas,v. 7, n. 1, p. 18--30, 2017}
\\
\newline
\\
\\
\\\hspace*{-9cm}\rule{50em}{0.2ex}

\title{On a method for Rock Classification using Textural Features and Genetic Optimization\\
{\footnotesize \it Sobre um m{\'e}todo de classifica\c{c}\~{a}o de rochas usando features de texturas e otimiza\c{c}\~{a}o gen{\'e}tica}}

\author{Manuel Blanco Valent{\'i}n}
\email{mbvalentin@cbpf.br}

\affiliation{Coordena\c{c}\~{a}o de Atividades T\'{e}cnicas (CAT/CBPF),\\
Centro Brasileiro de Pesquisas F\'{i}sicas\\
Rua Dr. Xavier Sigaud, 150, Ed. C\'{e}sar Lattes,\\
Urca, Rio de Janeiro, RJ. CEP 22290-180, Brasil}

\author{Cl{\'e}cio Roque de Bom}
\email{debom@cbpf.br}

\affiliation{Centro Federal de Educa\c{c}\~{a}o Tecnol\'{o}gica Celso Suckow da Fonseca,\\
Rodovia M\'{a}rio Covas, lote J2, quadra J\\
Distrito Industrial de Itagua\'{i},\\
Itagua\'{i} - RJ. CEP: 23810-000, Brasil}

\author{M{\'a}rcio P. de Albuquerque}
\email{mpa@cbpf.br}

\author{Marcelo P. de Albuquerque}
\email{marcelo@cbpf.br}

\author{Elis\^{a}ngela L. Faria}
\email{elisangela@cbpf.br}

\affiliation{Coordena\c{c}\~{a}o de Atividades T\'{e}cnicas (CAT/CBPF),\\
Centro Brasileiro de Pesquisas F\'{i}sicas\\
Rua Dr. Xavier Sigaud, 150, Ed. C\'{e}sar Lattes,\\
Urca, Rio de Janeiro, RJ. CEP: 22290-180, Brasil}

\author{Maury D. Correia}
\email{maury.duarte@petrobras.com.br}
\author{Rodrigo Surmas}
\email{surmas@petrobras.com.br}
\affiliation{Centro de Pesquisas e Desenvolvimento Leopoldo Am\'{e}rico Miguez de Mello -- CENPES\\
PETROBRAS, Av. Hor\'{a}cio Macedo, 950, Cidade Universit\'{a}ria,\\
 Rio de Janeiro, RJ. CEP 21941-915, Brasil\\
 \footnotesize  Submetido: 01/01/2016\ \ \ \ \ \ \ \ Aceito: 16/05/2016}

\begin{abstract}
\noindent
{\bf Abstract:} In this work we present a method to classify a set of rock textures based on a Spectral Analysis and the extraction of the texture Features of the resulted images. Up to 520 features were tested using 4 different filters and all $31$ different combinations were verified. The classification process relies on a Na\"{i}ve Bayes classifier. We performed two kinds of optimizations: statistical optimization with covariance-based Principal Component Analysis (PCA) and a genetic optimization, for 10,000 randomly defined samples, achieving a final maximum classification success of 91\% against the original $\sim$ 70\% success ratio (without any optimization nor filters used). After the optimization $9$ types of features emerged as most relevant. 

\vspace*{3mm}

\vspace*{3mm}
\noindent
{\bf Resumo:} Neste trabalho apresentamos um m\'{e}todo para classificar um conjunto de texturas de rocha baseado na An\'{a}lise espectral e na extra\c{c}\~{a}o de features texturais das imagens resultantes. Um conjunto de 520 features foi testado usando 4 filtros diferentes e todas as 31 combina\c{c}\~{o}es dos mesmos foram verificadas. O processo de classifica\c{c}\~{a}o proposto \'{e} baseado em um classificador Na\"{i}ve Bayes. Foram realizados dois tipos de otimiza\c{c}\~{a}o nos par\^{a}metros extra\'{i}dos: uma otimiza\c{c}\~{a}o estat\'{i}stica usando uma An\'{a}lise de Componente Principal por covari\^{a}ncia (PCA) e uma otimiza\c{c}\~{a}o gen\'{e}tica, para todas as 10.000 permuta\c{c}\~{o}es aleat\'{o}rias das imagens, obtendo um sucesso m\'{a}ximo final de classifica\c{c}\~{a}o de 91\%, sendo o sucesso inicial, sem nenhum tipo de otimiza\c{c}\~{a}o do 70\%. Depois da aplica\c{c}\~{a}o do m\'{e}todo aqui descrito 9 tipos diferentes de features emergeram como as mais relevantes para o problema de classifica\c{c}\~{a}o de texturas de rochas.

\vspace*{3mm}
\noindent
{\bf Palavras chave:} Par\^{a}metros de Haralick, Algoritmo Gen\'{e}tico, Classifica\c{c}\~{a}o de Texturas, Na\"{i}ve Bayes, Processamento de Imagens, An\'{a}lise espectral, PCA. 
\end{abstract}

\maketitle
\setcounter{page}{18}


%
%
\section{Introduction}
\label{sec1}

One of the most basic and important techniques on which we humans rely on our daily basis is image processing. Our brains and eyes have evolved in such a way that we can easily process the incoming images and make any decision very quickly. Distinguishing and segmenting the different textures contained in these images, as well as classifying them, are trivial tasks for almost any human being. \par

Nowadays, we use computers and artificial intelligence in several fields of science for multiple purposes, such as finding tumors in early stages to increase life expectancy of the pacient (see \cite{dhawan1986enhancement} and \cite{li1995fuzzy}), face recognition (see \cite{liu2002gabor} and \cite{ding2016multi}) or automatic classification of galaxies and stars (see \cite{kim2016star}). \par

In this context, rock classification can be very interesting in different areas of science, from geology to petrophysics. It has been long known that certain kinds of rock may produce more oil or gas than others. This is the reason why oil companies use different probes to gather information about the oil field in order to estimate the probability, presence and quantity of oil or gas on that certain field. The analysis of well logs has been relied over the years as a very powerful tool to aid analysts on deciding whether a field is suitable for exploration or not (see \cite{singh2016modelling}). \par

A part from well-log analysis, well images are also being used to identify patterns and formations in the well structure that may also complement the information extracted from the log curves. Acoustic and Electrical resistivity probes are commonly used for these purposes and the analysis and processing of these images can allow geologists to carry on a lithology study of the field, classifying the different types of rock in the walls of the drilled well and, therefore, gathering more information to make further solid decisions. \par

In this work we address the problem of rock texture classification by using Haralick Textural Features, along with other textural features, extracted from each image and a Na\"{i}ve Bayes classifier. In order to evaluate the process two texture datasets have been used: a standard Texture Dataset (KTH-TIPS), as a fiducial and well-known dataset used in these kinds of tests; and a Rock Texture Database (KCIMR - CENPES Rock Database), containing several samples of 9 different Rock classes. The features were evaluated in the original images and filtered images. We optimize our results with two approaches: by a Principal Component Analysis (PCA) and by Genetic Algorithm.\par 




This paper is organized as follows: In section 2 the process of textural features extraction is introduced. Section 3 briefly reviews the Na\"{i}ve Bayes classifier and how it can be used to classify textural features. In Section 4 the Spectral Analysis is presented along with the proposed filters. In Section 5 the main concept of Genetic Algorithms and its utility to optimize any set of textural features considered in classification is explained. The workflow used to classify the data is presented in Section 6, while the datasets used in this paper are shown in Section 7. Section 8 presents the classification results for all considered cases (with and without optimizations). Lastly, in Section 9 the conclusions achieved in this work are exposed.\par

%
%
\section{Rock Textural Classification Methodology}

\subsection{Textural Features}
\label{sec2}

Even though each texture classification problem is unique and will demand its own requirements and analysis, the classification method used for the task is, usually, straight-forward. Just like we humans do, computers classify objects (images or signals) by extracting visual patterns that may help characterize these objects, in such a way that the different classes or groups to be classified become distinguishable one from each other. \par

When treating images, these extracted visual patterns are called Textural features. Therefore, as introduced in the previous paragraph, it is expected -and desired- that each class or group will have very distinctive features, so that the classification task is easier. Usually, the optimal group of features that make the analyzed classes most distinguishable from each other is not known at the beginning, so several analysis have to be made in order to separate useful textural features from features that confuse the classifier (this is what is achieved in this paper by using PCA and Genetic Optimization).\par

In this paper we use 13 of the 14 textural features proposed by Haralick et al. in their original paper (see \cite{Ref12}). Haralick features are calculated using the so-called Gray-Level Co-occurrence matrix, which could be defined as a 2D distribution matrix that represents the probability of occurrence of a certain pair of graylevels in the image, given a certain offset (defining the neighborhood of this pair) and a certain direction. \par

These textural features have been widely used for pattern recognition and image classification since first published in 1973 with fairly good results (see e.g., \cite{PR4,PR5,Ref13,blanco2016texture, NTLuciana}). They belong to a class of textural features known as rotation-variant textural features. This means that the values of the extracted features depend on the orientation of the images, which is usually a not desired quality.\par 

Haralick et al. themselves proposed, in their original paper, to calculate the average values of the features in all possible directions for a given offset \footnote{As the images used here have reduced size (200x200 pixels), the offsets for the GLCM calculation will have a value of one pixel. On the other side, the selected number of graylevels has been 64.}. In this work all four directions (0\degree, 45\degree, 90\degree and 135\degree) have been considered, as well as their average values and their range values.\par

On the other hand, Linek et al. \cite{Ref11} proposed new features based on the co-occurrence matrix, which were then used to find patterns in resistivity borehole images to classify the rocks in the wall of the drilled borehole. In their paper they showed that these features seemed to be useful in borehole image classification, so they were included and used in this work (see \cite{NT}). These features are: Maximum Probability, Cluster Shade and Cluster Prominence.\par

Apart from these features, three extra textural features were considered for test in this paper: Tsallis Entropy  \cite{Ref17,tsallisAlbuquerque}), Fractal dimension (e.g., \cite{Ref18,Ref19}) and a Sato's Maximum Lyapunov Exponent (see \cite{sato1987practical}). \par

Thus, considering all parameters shown in this section, a total of 104 parameters will be obtained for each image\footnote{All 13 Haralick features plus 4 extra features for each GLCM, by 6 different GLCMs (4 offsets, Average and Range values to avoid anisotropy), along with the Fractal Dimension and the modified Lyapunvov Exponent.}. These parameters, after extracted from the original images, will be used as input data for the classifier.

\subsection{Na\"{i}ve Bayes Classifier}
\label{Naive Bayes Classifier}

There exist several approaches to classify data using a set of features. Among them, statistical classifiers are the most usual. This family of classifier based their operation, basically, in the computation of a certain cost function. Roughly speaking, the cost of each feature, of each sample to belong to each one of the possible classes is calculated; then, the class that showed the lowest cost is, commonly, the class predicted for that certain sample. \par

The classifier used in this paper is a Gaussian Na\"{i}ve classifier. Na\"{i}ve Bayes classifiers are based on the Bayes Theorem and the estimation of the posterior probability. According to this theorem, the probability of a certain set $X = \left(x_{1},x_{2}, \ldots ,x_{n} \right)$ to belong to a certain class $C_{k}$ 
is proportional to the product of individual probabilities for each feature to belong to that certain class. The decision rule, most times, is to simply assign the data $X$ to the class that obtained the greatest probability or, i.e., the class which had the highest value for the product of individual features probabilities. This decision rule is shown in (\ref{eq:5}).  the classifier used in this work is a Gaussian Na\"{i}ve Bayes.\par

\begin{equation}\label{eq:5}
    CLASS = \arg \max _{k=1...K} p(C_{k}) \prod_{i=1}^{n}{p(x_{i} | C_{k})}
\end{equation}

These classifiers assume that all variables are independent. Even though for cases where properties are dependent, several authors have shown that Na\"{i}ve Bayes stills reliable \cite{Ref23,Ref24}. A comparison between different types of classification methods (Na\"{i}ve Bayes included) can be found in \cite{keogh1999learning}.

\subsection{Spectral Analysis}
\label{Spectral Analysis}

As explained in Section \ref{sec2}, different textures are expected to have different textural features. The more distinctive these features are between the different classes, the more distinguishable the classes are and, therefore, the lesser misclassifications are expected to happen.\par

The truth is that textures, like in the ones considered in this case, are not necessarily distinguishable enough one from each other. One way to increase the distinguishability of classes is to filter these textures with different filters. This technique is not new, however it has been proved over the years that it actually helps increase the classification success ratio in these types of classification problems (see \cite{Filts1}). \par

The idea behind this technique is that when an image is filtered, its spectrum is changed and, obviously, so it is itself. This filtering process enhances certain parts of the spectrum of the original images, while it attenuates other aspects. Different textures (or classes of textures) may have similar textural features, however, they may have very different response to these filtering processes, leading to the extraction of new images from which new textural features -- features that can make classes more distinguishable -- can be extracted. \par

The filters tested in this paper are very common, however -- as we will show later in this article -- effective to increase the classification success ratio for this specific Rock classification problem. These are: a Low-pass Gaussian filter, an edge detector Canny filter\footnote{A more precise definition and explanation about the Canny filter can be read in John Canny$'$s original article \cite{Ref25}) and also in a later review  \cite{Ref26}.}, a 9-by-9 neighbor-box entropy filter and a 3-by-3 neighbor-box variance filter. \par

\begin{figure}[!htb]
    \centering
    \includegraphics[width=\linewidth]{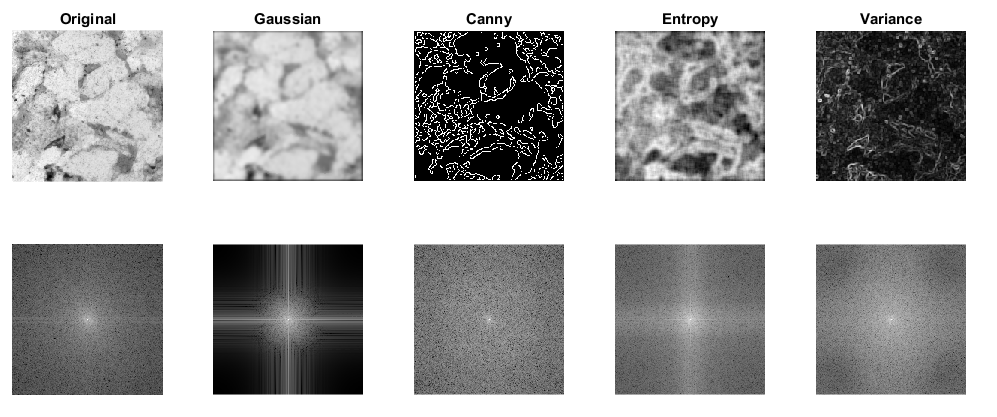}
    \caption{Effect of filtering in a sample of Buff Berea Sandstone from the KCIMR - CENPES Rock Database in $4$ filters. The second row shows the spectrum of the above image.}
    \label{fig:fig2}
\end{figure}

Fig. \ref{fig:fig2} shows an example of application of these filters in a sample of Buff Berea Sandstone. It can be seen how the application of the filters modifies the spectrum of the original image (at the left). Different filters produce different filtered images of the same sample, allowing us to increase our feature space and, therefore, icreasing the chances of finding the optimal subset of features.\par

%
%
\subsection{Principal Component Analysis and Genetic Optimization}
\label{Genetic Optimization}

A large number of features does not necessarily imply an improvement in the classifier performance and some of these features can be very similar between different classes for a certain classification problem, as introduced in Section \ref{sec2}. If that occurs it is necessary to separate the useful features that make the problem classes more distinguishable from those that cause the opposite.\par

Different methods have been proposed over the last decades to find an optimal subset within a group of features (see \cite{PR1,PR2}), in order, for instance, to increase the classification success -- like in this paper. When the features space is reduced (up to around 20 features), it might still be plausible to test all different possible combinations of features in order to find the one that achieves the best classification result. \par

Otherwise, when the features set is too large this solution becomes unpractical. In this paper, for instance, we use up to 520 features in some cases (when all filtered images are used, see Section \ref{Spectral Analysis}), which means that exist $3\cdot10^{+156}$ different combinations to be tested. In most applications testing all these combinations would be impossible due to computational costs. \par

The most common approaches to find the optimal subset of features are based on whether on statistical analysis of the features set or in stochastic brute-force algorithms (see a comparison of techniques in \cite{kohavi1997wrappers}). In this work we use one of each method to optimize our feature space in order to increase classification success.\par

The statistical analysis implemented in this paper is the widely used Covariance Principal Component Analysis. This analysis consists on simply obtaining the covariances coefficients between all different feature vectors. These coefficients can be then used to find a new normalized uncorrelated and independent set of features. The theory is that every set of vectors (features) can be redefined as a composition of a certain number of principal components. \par

These principal components are the new set of features. In order to reduce the feature space dimensionality only a fraction of these new features are considered. In this paper, for instance, only the first features that, after the PCA, gathered 95\% of the total variance of the optimized feature set were considered as optimal, while the rest of the feature space was discarded. \par

Although this technique reduces the problem dimensionality it does not guarantee that the optimal subset of features will be found. On the other side, Genetic Algorithms (GA) (see, e.g., \cite{Ref27}) are very useful tools that help the user to find local maxima or minima faster than classical optimization algorithms.\par

These algorithms belong to the search heuristic methods family. Their method mimics the genetics evolutionary theory, by evaluating the success of every individual, discarding the ones which had less success, mixing randomly the most successful ones to create new generations of individuals and introducing random mutations, until global maximum is achieved. In this paper the Genetic optimization process only stopped when the change in the average classification success from one iteration to the next one was below 0.001\%.\par

Thus, in this paper we have used, first, a PCA optimization to reduce feature dimensionality and, later, a Genetic Optimization to find the optimal subset of features that maximize classification success. \par

\section{Texture Classification Algorithm}
\label{Texture Classification}

%
%
Previous to the classification procedure, all original images are imported along with their respective rock class. Then they are all filtered using the four filters introduced in Section \ref{Spectral Analysis}. After that the textural features introduced in Section \ref{sec2} are extracted for all five images for each sample (Original + Gaussian-filtered + Canny-filtered + Entropy-filtered + Variance-filtered Images). \par

Once these features have been extracted we end up with 520 values of features for each sample, separated into 5 groups of 104, each group regarding each source of image (again, Original, Gaussian, Canny, Entropy or Variance). This division will later allow us to pick only the features of a certain filter for all samples, so that the utility of that filter in the classification problem is tested. \par

Thus, 31 different combinations will be tested (different possible combinations of 5 types of images/filters to use). This means that, for instance, combination 1 will only use the features extracted from the original images, while combination 31 will use the features extracted from all 5 images (Original + 4 filters).\par

All tests shown in this paper used 60\% of the sample set to train the classifier and the rest of the samples to test it.\par

For each tested combination three classification tests are carried out: First the features of that certain combination of images, without any type of optimization; second, the same subset of features is optimized using only PCA before using it to classify; and third, the same subset of features is optimized by PCA and then by a Genetic Algorithm. Step one will gives us ground-control data that we can use to compare to the results further obtained after PCA and Genetic Optimization in order to evaluate if these methods do actually improve classification success. \par

In each one of these steps 10,000 different iterations are tested to obtain more robust and reliable results. Each iteration sorts the data randomly, in an attempt to remove any possible relation between a choice of a particular samples and the classification success. After this permutation is done, the data is split into training (60\%) and testing group (40\%). The first one will be used to train a gaussian Na\"{i}ve Bayes classifier. This same classifier will then be fed with the second group (testing), producing a prediction class for each one of the testing samples. This predictions can be then compared with the real classes of the testing data subset, obtaining the classification success ratio. \par

\begin{figure*}[!t]
    \centering
    \includegraphics[width=\textwidth,keepaspectratio]{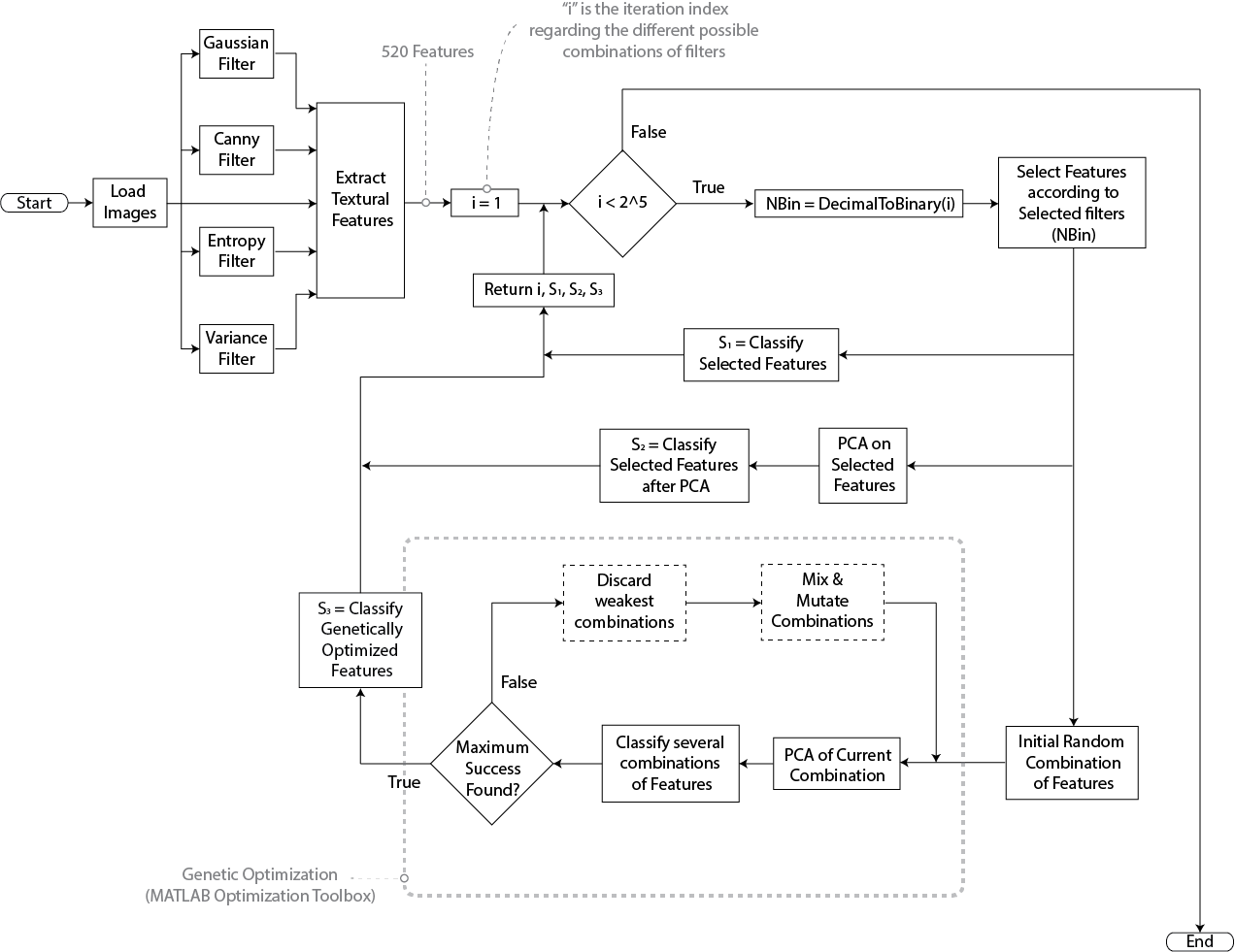}
    \caption{Organigram of the algorithm used in this work to classify the data.}
    \label{fig:organn}
\end{figure*}

A diagram of the algorithm used in this work for the classification process is illustrated on Fig. \ref{fig:organn}.



%
%
\subsection{About the KTH-TIPS Dataset}
\label{About the KTH-TIPS Dataset}
The first dataset of images used as samples for training and testing our classifier is called KTH-TIPS and was firstly used by Hayman et al. \cite{Ref32} and shortly after that became available for public use. Since then this library of images has been widely used,  as examples of textures for image processing, analyzing, filtering and classification (e.g., \cite{Ref33,Ref34}).\par

This dataset provides a total of 810 images, divided in 10 different classes. A more extensive description of this database can be found in \cite{Ref36}. The materials, and therefore the classes, found in this dataset are: \par

\begin{enumerate}
    \item Sandpaper (SD)
    \item Aluminum Foil (AL)
    \item Styrofoam (SY)
    \item Sponge (SP)
    \item Corduroy (CY)
    \item Linen (LI)
    \item Cotton (CT)
    \item Brown Bread (BB)
    \item Orange Peel (OP)
    \item Cracker Biscuit (CR)
\end{enumerate}

%
%
\subsection{About the KCIMR - CENPES Rock Database}
\label{About the KCIMR-CENPES Rock Dataset}
The second dataset of images used as samples for training and testing our classifier is the KCIMR - CENPES Rock Database\footnote{Kocurek Carbonate, Igneous and Mineral Rocks - CENPES Rock Dataset. Classes BBS, DPL, EYC, GBS, IBS, IL and SD are carbonates, while GNT is Igneous and OLI is a mineral.}. This dataset is the result of the combination of three different datasets of rock textures, one produced by the authors and the other two are public avaliable. All images were obtained by optical microscope. The Kocurek Carbonates Dataset, which represents the different thin section of plugs of the BBS, DPL, EYC, GBS, IBS, IL and SD carbonate rock classes, produced in CENPES\footnote{Centro de Pesquisas Leopoldo Am\'{e}rico Miguez de Mello} Laboratory by the CENPES Tomography group led by R. Surmas; Granite sample images from the GeoSecSlides group\footnote{\url{http://www.geosecslides.co.uk/}}; and a group of Olivinite sample images from the NCPTT\footnote{National Center for Preservation Technology and Training.} of the National Park Service public images\footnote{\url{http://ncptt.nps.gov/buildingstone/stone/adirondack-granite}}.\par

\begin{figure*}[!htb]
    \centering
    \includegraphics[width=\textwidth]{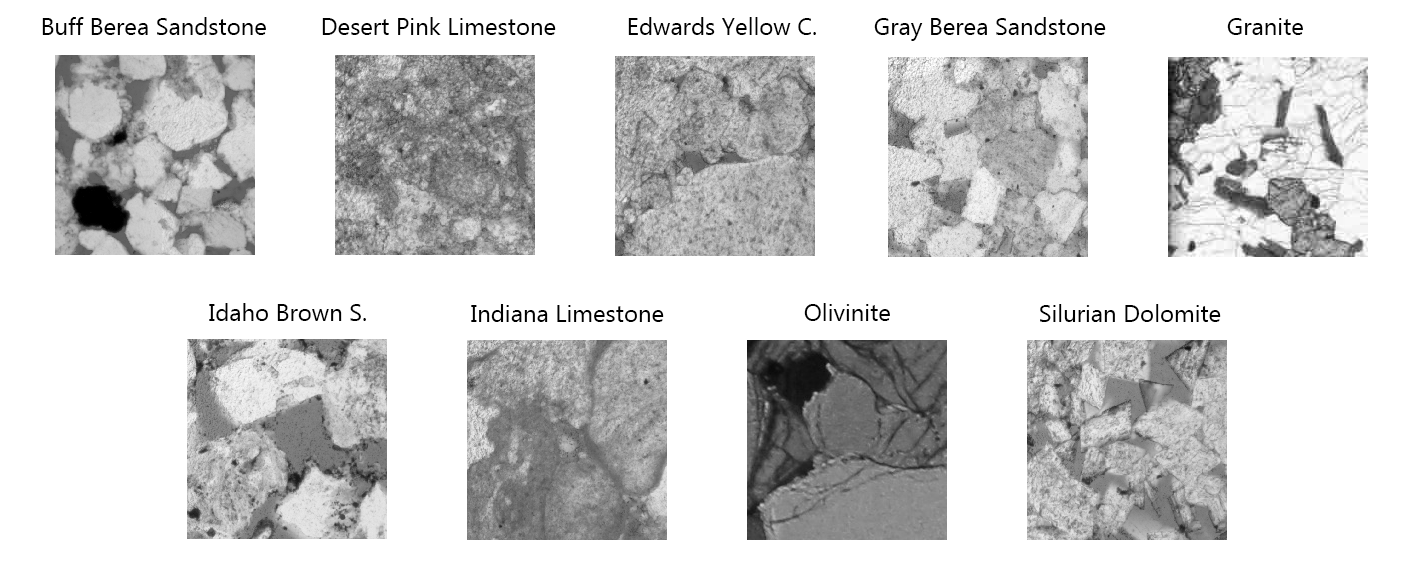}
    \caption{Sample images from all 9 different classes in the Rock dataset.}
    \label{fig:fig.4}
\end{figure*}

This dataset provides a total of 2,520 images, divided in 9 different classes with 280 pictures for each class. A sample of these textures can be seen in Fig. \ref{fig:fig.4}. The materials, and therefore the classes, found in this dataset are: \par
\begin{enumerate}
    \item Buff Berea Sandstone (BBS)
    \item Desert Pink Limestone (DPL)
    \item Edwards Yellow Carbonate (EYC)
    \item Gray Berea Sandstone (GBS)
    \item Granite (GNT)
    \item Idaho Brown Sandstone (IBS)
    \item Indiana Limestone (IL)
    \item Olivinite (OLI)
    \item Silurian Dolomite (SD)
\end{enumerate}

The carbonate image classes, such as the ones obtained for this dataset, are particularly relevant to oil and gas industries with some applications mentioned in section \ref{sec1}. It is worth saying that, besides the classification of these textures may not be the fully representative when it comes to other type of rock images in different scales, such as acoustic and resistivity patterns, they can be very useful to test and improve algorithms and methods related to rock classification, and to give insights into the data, regardless of the geological classification of these rock textures.\par

%
%
\section{Classification Results}
\label{Classification Results}

The classification results for a training set of 60\% and a testing set of 40\%, for the KCIMR - CENPES Rock Dataset images for all filters combinations (31 cases) is shown in Table \ref{tab:resultsRock}, while the results for the KTH-TIPS Dataset for all combinations is shown in Table \ref{tab:resultsKTH}.\par

As it can be seen in these two tables, the average classification success when no filters nor optimization processes were used was $(70.20 \pm 1.31)\%$ for the KCIMR database and $(71.96 \pm 2.26)\%$ for the KTH-TIPS database.\par

The classification rate values (count of times that a certain real class was classified as another class, in average) is shown in Table \ref{tab:miscKCIMR} for the KCIMR Database and in Table \ref{tab:miscKTH} for the KTH--TIPS Database. From these tables one may infer how well defined a class is or which are the most commonly misclassified classes. In the next subsection we discuss the impact of each test we performed.\par


\subsection{Impact of Spectral Analysis on Classification}
\label{Impact of Spectral Analysis on Classification}
We evaluate the correlation between the filters and the classification success for the original images, Variance filter, Entropy filter, Canny filter and Gaussian filter. The results were $45.82\%$, $26.25\%$, $-25.47\%$, $26.64\%$, $37.43\%$  for KTH-TIPS Dataset and $38.82\%$, $32.43\%$, $-27.36\%$, $26.62\%$ \& $50.78\%$ for KCIMR Dataset respectively. Even though the used texture datasets are significantly different one from each other, both cases showed positive results when using most of filters, except for Entropy Filter. Therefore, this filter should not be used in further tests using any of the two datasets analyzed in this paper. The Gaussian filter present to be particularly valuable for KCIMR Dataset.\par

The maximum success configuration (Case 23) due to other $3$ filters was 10.58\% for the KCIMR database and 8.73\% for the KTH-TIPS database. A comparison between the classification results before and after the filtering process, for both datasets, is shown in Table \ref{tab:impact1}.\par  






\begin{table}[!htb]
\centering
  \begin{tabular}{| c | c | c |}
    \hline
    Dataset & Original Image Sucess & Sucess After SA\\
    \hline
    KCIMR & $(70.20 \pm 1.31)\%$ & $(80.78 \pm 1.05)\%$\\
    KTH-TIPS & $(71.96 \pm 2.26)\%$ & $(80.69 \pm 1.86)\%$\\
    \hline
  \end{tabular}
\caption{Impact of SA on Classification}
\label{tab:impact1}
\end{table}

\subsection{Impact of PCA on Classification}
\label{Impact of PCA on Classification}
The second technique considered to improve the success ratio was the Principal Component Analysis. The classification success ratio increased in all tested cases. The maximum increase on success due to the PCA only (i.e. diference before and after PCA for each case) was 12.85\% for the KCIMR database and 12.52\% for the KTH-TIPS database, both for case 17. A comparison between the classification results before and after the PCA optimization process, for both datasets, in the best filter configuration, case 23, is shown in Table \ref{tab:impact2}.\par  
\begin{table}[!htb]
\centering
\caption{Impact of PCA on Classification}
\label{tab:impact2}
  \begin{tabular}{| c | c | c |}
    \hline
    Dataset & Original Images Success & Success after SA+PCA\\
    \hline
    KCIMR & $(70.20 \pm 1.31)\%$ & $(88.01 \pm 0.94)\%$\\
    KTH-TIPS & $(71.96 \pm 2.26)\%$ & $(86.36 \pm 1.85)\%$\\
    \hline
  \end{tabular}
\end{table}

\subsection{Impact of Genetic Optimization on Classification}
\label{Impact of Genetic Optimization on Classification}
 When this optimization method the maximum increase on success due to the Genetic Optimization was 19.08\% (i.e. diference before and after GA with embed PCA for each case) for the KCIMR database and 16.11\% for the KTH-TIPS database, in case 16. The standard deviation value of the classification success ratio was reduced, on average, 0.15\% for the KCIMR database and 0.28\% for the KTH-TIPS database. A comparison between the best classification results (Case 23), regarding all $3$ optimizations, for both datasets, is shown in Table \ref{tab:impact3}.\par  
 
\begin{table}[!htb]
\centering
\caption{Impact of Genetic optimization on Classification}
\label{tab:impact3}
  \begin{tabular}{| c | c | c |}
    \hline
    Dataset & Original Images & After SA+PCA+GA\\
    \hline
    KCIMR & $(70.20 \pm 1.31)\%$ & $(91.15 \pm 0.86)\%$\\
    KTH-TIPS & $(71.96 \pm 2.26)\%$ & $(92.27 \pm 1.59)\%$\\
    \hline
  \end{tabular}
\end{table}


\subsubsection{Most relevant features}
\label{Most relevant features}

It is worth noticing that this optimization process can reduce the number of features used. The Genetic Optimization reduced the number of features approximately in half, on average, for both cases. In this case, the features that were mostly preserved after the optimization (and, therefore, the features that can be considered the optimal for classifying these textures) were (for both datasets):




\begin{enumerate}
    \item Entropy\footnote{\label{fnt:hhnote}These features belong to the original Haralick Features set, see \cite{Ref12}.}
    \item Diff. Variance\footnotemark[\getrefnumber{fnt:hhnote}].
    \item Diff. Entropy\footnotemark[\getrefnumber{fnt:hhnote}]. 
    \item Cluster Shade\footnote{\label{fnt:f2note}These features belong to the features proposed by M. Linek et al., see \cite{Ref11}.}
    \item Cluster Prominence\footnotemark[\getrefnumber{fnt:f2note}]. 
    \item Correlation\footnotemark[\getrefnumber{fnt:f2note}].
    \item Local Homogeneity\footnotemark[\getrefnumber{fnt:f2note}].
\end{enumerate}

Also, two of the three extra features proposed in this paper (see \ref{sec2}). the Fractal Dimension and the MLE values for each image were optimal and used to improve classification success. The Tsallis Entropy values were used as often as any other of the Haralick features not considered in the previous list.\par

\subsection{Best results comparison}
\label{Best results comparison}
The best three results obtained for each one of the datasets after the optimization and filtering processes are shown -- along with the original images case -- in Table \ref{tab:top3} and Table \ref{tab:top3kth}.\par

As it can be seen in Table \ref{tab:top3}, all three best results have very similar values. Roughly, the classification success was increased up to 20\% with the optimization and filtering process. Even though case 23 achieved the best result, cases 19 and 21 require fewer features to be extracted and analyzed from every single image. This statement is also true for the KTH--TIPS Database, as it can be seen in Table \ref{tab:top3kth}. In this case, the classification success also increased up to 21\%, approximately although the case $7$ emerges as third option instead of case $21$ in the rock texture sample.\par

For any case, the choice of the best case will depend on the requirements of each single application.\par




According to \ref{tab:miscKCIMR} the most common misclassifications in KCIMR - CENPES  occur mainly between classes SD and EYC, and then between classes OLI and GNT or GBS and IBS. For KTH--TIPS Database, Table \ref{tab:miscKTH} suggests that the most common misclassifications occur mainly between classes LI and CT, and then between classes CY and CT or OP and SY.


\begin{table}[!htb]
\centering
\caption{Results comparison for KCIMR -- CENPES Rock Dataset.}
\label{tab:top3}
  \begin{tabular}{| c | c | c | c |}
    \hline
    Case & Avg. Success. & $NF_{GA}$ & Filters\\
    \toprule
    1 & $70.20\%$ & 104 (104) & Original images only.\\
    23 & $91.15\%$ & 180 (416) & No Entropy  \\
    19 & $90.29\%$ & 137 (312) & No Canny and Entropy \\
    21 & $90.12\%$ & 141 (312) & No Gaussian and Entropy \\
    \hline
  \end{tabular}
\end{table}

\begin{table}[!htb]
\centering
\caption{Results comparison for KTH--TIPS Dataset.}
\label{tab:top3kth}
  \begin{tabular}{| c | c | c | c |}
    \hline
    Case & Avg. Success. & $NF_{GA}$ & Filters\\
    \hline
    1 & $71.96\%$ & 104 (104) & Original images only\\
    23 & $92.27\%$ & 202 (416) & No Entropy \\
    19 & $91.91\%$ & 154 (312) & No Canny and Entropy \\
    7 & $91.61\%$ & 142 (312) & No Variance and Entropy \\
    \hline
  \end{tabular}
\end{table}

\newpage
%
%
\section{Conclusions}
\label{Conclusions}
In this work we have proposed a workflow to increase the classification success ratio in Na\"{i}ve bayes classifiers by using image filters, principal component analysis and genetic optimization algorithms and exhaustively tested up to $520$ features for rock texture classification applications. 

We apply this approach in two different sets of samples: a well known and widely used texture database (KTH--TIPS) and a rock texture database -- described in this work which its major part was produced to test the proposed algorithm -- used to address the question of the viability of rock textures classification, in particular carbonate textures which are of extreme interest to oil and gas industries. The results shown in the previous sections allow us to conclude that: 

\begin{enumerate}
    \item The Spectral Analysis shown in this paper, that $3$ out of $4$ filters tested the Gaussian, Canny and Variance filtered images along with the original ones showed positive results, increasing notably the classification success ratio up to 10\% (for the KCIMR - CENPES Rock Database), suggesting that they  can be useful to enhance some of the features that are hidden in the original images, improving the classification success with little effort and computational cost.
    
    \item The Principal Component Analysis showed significant positive results when it comes to improving the classification success. When this technique was applied to the features the classification success was increased up to $ \sim 13\%$ (for the KCIMR - CENPES Rock Database). 
    \item The Genetic Optimization used in this work also allowed us to increase our classifier success ratio some points up. The combination of three types of optimization improved this success up to 19\% (for the KCIMR - CENPES Rock Database). This optimization allowed the classifier to reach a classification success ratio above 91\%, for both datasets.
    \item The number of features after the genetic optimization process was reduced, in average, to half the original number of features.
    \item For some cases, some of the 10,000 permutations presented a very high classification success ratio. For instance, when analyzing the KTH--TIPS dataset, two cases showed an absolute maximum classification success ratio value over 97\%; while for the KCIMR dataset two permutations had this value over 93.5\%.
    \item After the combined filtering and optimization processes shown in this paper not only the classification success ratio increased substantially, but also the standard deviation of this ratio (for the 10,000 different random permutations) decreased. This parameter went from 1.31\par to 0.86\% for the best case of the KCIMR - CENPES Rock Database, while it went from 2.26\% to 1.59\% for the best case of the KTH--TIPS Dataset. 
    \item After the Genetic Optimization $9$ classes of features emerged as the most relevant for the classification of the tested textures. One of them, to the best of our knowledge, has never been proposed as a texture feature: the MLE.
\end{enumerate}

As shown in this paper this workflow allows the user to improve significantly the classification success ratio for any textural data. In both datasets studied here this ratio was increased from 70\% to over 91\%.\par

On the other hand, the implementation of the rock classification workflow with more sophisticated approaches, like Neural Networks, random forests for example has not been fully tested in our rock dataset. This is currently under investigation.

\section*{Acknowledgments}
This work was made possible by cooperation agreement between CENPES/PETROBR{\'A}S and CBPF and was supported by CARMOD thematic funding for Researches in Carbonates. C.R. Bom would aldo like to thank CNPq.

\bibliographystyle{unsrt}
\bibliography{refs}

\newpage
\onecolumngrid
\appendix
\newpage
\section{Test Results}\label{App:AppendixA}

\bgroup
\setlength\tabcolsep{8pt}
\setlength{\extrarowheight}{5pt}
\begin{table}[!ht]
\centering
\caption{Classification Results for the KTH--TIPS Dataset.}
\label{tab:resultsKTH} 
      \begin{tabular}{| c | c c c c c | c c | c c | c c | c c |}
      \hline
        & V & E & C & G & O & $\mu_{0}^{\%}$ & $\sigma_{0}^{\%}$ & $\mu_{PCA}^{\%}$ & $\sigma_{PCA}^{\%}$ & $\mu_{GA}^{\%}$ & $\sigma_{GA}^{\%}$ & $NF_{0}$ & $NF_{GA}$ \\
      \hline
      1 & 0 & 0 & 0 & 0 & 1 & 71.96 & 2.26 & 83.05 & 2.05 & 86.54 & 1.89 & 104 & 55 \\
        2 & 0 & 0 & 0 & 1 & 0 & 71.59 & 2.23 & 70.43 & 2.42 & 83.39 & 2.00 & 104 & 45 \\
        3 & 0 & 0 & 0 & 1 & 1 & 78.77 & 2.05 & 83.68 & 1.96 & 90.74 & 1.77 & 208 & 114 \\
        4 & 0 & 0 & 1 & 0 & 0 & 51.18 & 2.30 & 55.57 & 2.29 & 60.24 & 2.25 & 104 & 54 \\
        5 & 0 & 0 & 1 & 0 & 1 & 76.67 & 2.08 & 85.78 & 1.87 & 89.26 & 1.74 & 208 & 122 \\
        6 & 0 & 0 & 1 & 1 & 0 & 77.21 & 2.07 & 75.98 & 2.24 & 90.27 & 1.67 & 208 & 100 \\
        7 & 0 & 0 & 1 & 1 & 1 & 80.49 & 1.88 & 85.04 & 1.98 & 91.61 & 1.72 & 312 & 142 \\
        8 & 0 & 1 & 0 & 0 & 0 & 20.42 & 1.96 & 24.05 & 2.04 & 20.58 & 2.86 & 104 & 57 \\
        9 & 0 & 1 & 0 & 0 & 1 & 72.90 & 2.23 & 73.41 & 2.44 & 83.92 & 1.97 & 208 & 98 \\
        10 & 0 & 1 & 0 & 1 & 0 & 72.16 & 2.20 & 59.55 & 2.50 & 77.47 & 2.12 & 208 & 83 \\
        11 & 0 & 1 & 0 & 1 & 1 & 79.16 & 2.05 & 79.61 & 2.21 & 88.82 & 1.75 & 312 & 142 \\
        12 & 0 & 1 & 1 & 0 & 0 & 51.95 & 2.32 & 42.94 & 2.76 & 28.33 & 5.77 & 208 & 96 \\
        13 & 0 & 1 & 1 & 0 & 1 & 76.90 & 2.11 & 81.68 & 2.13 & 87.70 & 1.76 & 312 & 155 \\
        14 & 0 & 1 & 1 & 1 & 0 & 77.62 & 2.02 & 71.49 & 2.35 & 78.89 & 2.18 & 312 & 137 \\
        15 & 0 & 1 & 1 & 1 & 1 & 80.70 & 1.84 & 83.00 & 2.03 & 88.26 & 1.86 & 416 & 195 \\
        16 & 1 & 0 & 0 & 0 & 0 & 64.93 & 2.46 & 77.45 & 2.08 & 78.07 & 2.04 & 104 & 51 \\
        17 & 1 & 0 & 0 & 0 & 1 & 74.01 & 2.40 & 85.12 & 1.93 & 90.12 & 1.87 & 208 & 108 \\
        18 & 1 & 0 & 0 & 1 & 0 & 78.32 & 1.98 & 82.62 & 1.91 & 87.77 & 1.87 & 208 & 95 \\
        19 & 1 & 0 & 0 & 1 & 1 & 79.97 & 1.98 & 85.59 & 1.86 & 91.91 & 1.77 & 312 & 154 \\
        20 & 1 & 0 & 1 & 0 & 0 & 69.69 & 2.29 & 81.05 & 2.06 & 83.87 & 1.95 & 208 & 112 \\
        21 & 1 & 0 & 1 & 0 & 1 & 76.58 & 2.09 & 86.25 & 1.90 & 90.90 & 1.69 & 312 & 157 \\
        22 & 1 & 0 & 1 & 1 & 0 & 79.39 & 1.93 & 83.81 & 1.93 & 89.72 & 1.78 & 312 & 158 \\
        23 & 1 & 0 & 1 & 1 & 1 & 80.69 & 1.86 & 86.36 & 1.85 & 92.27 & 1.59 & 416 & 202 \\
        24 & 1 & 1 & 0 & 0 & 0 & 65.61 & 2.44 & 68.61 & 2.60 & 78.53 & 2.09 & 208 & 107 \\
        25 & 1 & 1 & 0 & 0 & 1 & 74.48 & 2.40 & 82.84 & 2.03 & 86.64 & 1.75 & 312 & 140 \\
        26 & 1 & 1 & 0 & 1 & 0 & 78.70 & 1.98 & 78.59 & 2.08 & 88.99 & 1.86 & 312 & 145 \\
        27 & 1 & 1 & 0 & 1 & 1 & 80.27 & 1.96 & 84.06 & 1.93 & 91.50 & 1.57 & 416 & 189 \\
        28 & 1 & 1 & 1 & 0 & 0 & 70.07 & 2.29 & 76.57 & 2.20 & 81.70 & 2.06 & 312 & 159 \\
        29 & 1 & 1 & 1 & 0 & 1 & 76.74 & 2.10 & 84.33 & 1.99 & 90.01 & 1.65 & 416 & 189 \\
        30 & 1 & 1 & 1 & 1 & 0 & 79.67 & 1.92 & 81.55 & 2.04 & 87.19 & 1.78 & 416 & 210 \\
        31 & 1 & 1 & 1 & 1 & 1 & 80.85 & 1.84 & 85.08 & 1.91 & 90.45 & 1.66 & 520 & 237 \\
        \hline
    \multicolumn{14}{@{}l}{\footnotesize{Filters:  V.--Variance  /  E.--Entropy  /  C.--Canny  /  G.--Gaussian / O.--Original}}\\
    \multicolumn{14}{@{}l}{\footnotesize{$\mu_{0}$--Avg. Success (Original Images)\qquad$\sigma_{0}$--Success Std. Deviation (Original Images)}}\\
    \multicolumn{14}{@{}l}{\footnotesize{$\mu_{PCA}$--Average Success (after PCA)\qquad$\sigma_{PCA}$--Success Standard Deviation (after PCA)}}\\
    \multicolumn{14}{@{}l}{\footnotesize{$\mu_{GA}$--Average Success (after G.O)\qquad$\sigma_{GA}$--Success Standard Deviation (after G.O.)}}\\
    \multicolumn{14}{@{}l}{\footnotesize{$NF_{0}$--Initial Number of Features\qquad$NF_{GA}$--Number of Features (after G.O.)}}
    \end{tabular}
\end{table}
\egroup

\begin{table*}[!ht]
  \centering
  \caption{Average confusion matrix for classification results on Case 23 for KTH--TIPS Dataset (in absolute values, not percentage). Rows represent the real classes of the images used while columns represent the prediction result of the classifier. High values in the diagonal of this matrix represent true positives, while values outside the diagonal represent misclassifications.}
  \label{tab:miscKTH}
      \begin{tabular}{| p{0.05\textwidth} | p{0.06\textwidth} p{0.06\textwidth} p{0.06\textwidth} p{0.06\textwidth} p{0.06\textwidth} p{0.06\textwidth} p{0.06\textwidth} p{0.06\textwidth} p{0.06\textwidth} p{0.06\textwidth} |}
    	\hline
    	\, & CY & LI & CR & BB & OP & SY & CT & SD & AL & SP \\
    	\hline
        CY & 82.02 & 6.46 & 0.32 & 0.9 & 0.26 & 0.03 & 6.8 & 0.9 & 2.32 & 0 \\ 
        LI & 0 & 85.51 & 0 & 0.22 & 0.53 & 0 & 13.38 & 0.33 & 0.03 & 0 \\ 
        CR & 0 & 0 & 90.04 & 1.84 & 1.75 & 0.21 & 3.44 & 1.48 & 0 & 1.24 \\ 
        BB & 0 & 0 & 0 & 92.01 & 1.21 & 0.03 & 1.36 & 0.62 & 0 & 4.78 \\ 
        OP & 0 & 0 & 0 & 0 & 86.39 & 6.38 & 0.3 & 3.09 & 2.84 & 0.99 \\ 
        SY & 0 & 0 & 0 & 0 & 0 & 94.39 & 0.03 & 5.51 & 0 & 0.07 \\ 
        CT & 0 & 0 & 0 & 0 & 0 & 0 & 96.77 & 2.7 & 0.25 & 0.28 \\ 
        SD & 0 & 0 & 0 & 0 & 0 & 0 & 0 & 99.93 & 0 & 0.07 \\ 
        AL & 0 & 0 & 0 & 0 & 0 & 0 & 0 & 0 & 100 & 0 \\ 
        SP & 0 & 0 & 0 & 0 & 0 & 0 & 0 & 0 & 0 & 100 \\ 

        \hline
      \end{tabular}
\end{table*}

\bgroup
\setlength\tabcolsep{8pt}
\setlength{\extrarowheight}{5pt}
\begin{table}[!ht]
  \centering
  \caption{Classification Results for the KCIMR -- CENPES Rock Dataset}
  \label{tab:resultsRock} 
      \begin{tabular}{| c | c c c c c | c c | c c | c c | c c |}
      \hline
        & V & E & C & G & O & $\mu_{0}^{\%}$ & $\sigma_{0}^{\%}$ & $\mu_{PCA}^{\%}$ & $\sigma_{PCA}^{\%}$ & $\mu_{GA}^{\%}$ & $\sigma_{GA}^{\%}$ & $NF_{0}$ & $NF_{GA}$ \\
        \hline
        1 & 0 & 0 & 0 & 0 & 1 & 70.20 & 1.31 & 80.55 & 1.25 & 86.74 & 1.01 & 104 & 56 \\
        2 & 0 & 0 & 0 & 1 & 0 & 72.49 & 1.18 & 76.93 & 1.19 & 82.65 & 1.04 & 104 & 45 \\
        3 & 0 & 0 & 0 & 1 & 1 & 75.32 & 1.17 & 82.95 & 1.13 & 88.05 & 1.09 & 208 & 87 \\
        4 & 0 & 0 & 1 & 0 & 0 & 50.65 & 1.29 & 58.19 & 1.22 & 59.25 & 1.28 & 104 & 58 \\
        5 & 0 & 0 & 1 & 0 & 1 & 74.57 & 1.16 & 83.49 & 1.10 & 89.27 & 1.01 & 208 & 106 \\
        6 & 0 & 0 & 1 & 1 & 0 & 77.59 & 1.09 & 82.38 & 1.06 & 85.74 & 0.94 & 208 & 102 \\
        7 & 0 & 0 & 1 & 1 & 1 & 79.05 & 1.05 & 84.86 & 1.02 & 88.93 & 1.01 & 312 & 139 \\
        8 & 0 & 1 & 0 & 0 & 0 & 21.39 & 1.28 & 25.47 & 1.16 & 25.81 & 1.18 & 104 & 65 \\
        9 & 0 & 1 & 0 & 0 & 1 & 59.79 & 2.67 & 48.43 & 3.27 & 85.94 & 1.15 & 208 & 91 \\
        10 & 0 & 1 & 0 & 1 & 0 & 62.67 & 2.95 & 54.68 & 3.05 & 81.08 & 1.13 & 208 & 91 \\
        11 & 0 & 1 & 0 & 1 & 1 & 69.55 & 1.69 & 65.01 & 2.50 & 85.31 & 1.05 & 312 & 154 \\
        12 & 0 & 1 & 1 & 0 & 0 & 44.50 & 3.15 & 34.17 & 2.50 & 61.85 & 1.34 & 208 & 76 \\
        13 & 0 & 1 & 1 & 0 & 1 & 70.10 & 1.69 & 53.67 & 2.93 & 87.03 & 0.97 & 312 & 130 \\
        14 & 0 & 1 & 1 & 1 & 0 & 73.44 & 1.79 & 60.37 & 2.80 & 84.73 & 1.00 & 312 & 119 \\
        15 & 0 & 1 & 1 & 1 & 1 & 76.88 & 1.40 & 68.99 & 2.33 & 88.27 & 0.96 & 416 & 189 \\
        16 & 1 & 0 & 0 & 0 & 0 & 60.45 & 1.27 & 72.12 & 1.35 & 79.53 & 1.18 & 104 & 39 \\
        17 & 1 & 0 & 0 & 0 & 1 & 71.93 & 1.27 & 84.78 & 1.14 & 87.83 & 1.05 & 208 & 113 \\
        18 & 1 & 0 & 0 & 1 & 0 & 74.89 & 1.19 & 85.05 & 1.04 & 88.16 & 0.91 & 208 & 95 \\
        19 & 1 & 0 & 0 & 1 & 1 & 75.99 & 1.16 & 86.98 & 1.00 & 90.29 & 0.92 & 312 & 137 \\
        20 & 1 & 0 & 1 & 0 & 0 & 69.95 & 1.16 & 77.23 & 1.17 & 83.75 & 1.10 & 208 & 107 \\
        21 & 1 & 0 & 1 & 0 & 1 & 77.64 & 1.11 & 86.62 & 1.00 & 90.12 & 0.88 & 312 & 141 \\
        22 & 1 & 0 & 1 & 1 & 0 & 80.44 & 1.05 & 87.07 & 0.98 & 89.20 & 0.91 & 312 & 154 \\
        23 & 1 & 0 & 1 & 1 & 1 & 80.78 & 1.05 & 88.01 & 0.94 & 91.15 & 0.86 & 416 & 180 \\
        24 & 1 & 1 & 0 & 0 & 0 & 55.78 & 2.51 & 44.93 & 2.51 & 77.78 & 1.26 & 208 & 76 \\
        25 & 1 & 1 & 0 & 0 & 1 & 70.31 & 1.72 & 58.96 & 2.35 & 86.60 & 1.07 & 312 & 133 \\
        26 & 1 & 1 & 0 & 1 & 0 & 73.72 & 1.74 & 66.57 & 2.42 & 88.15 & 0.99 & 312 & 118 \\
        27 & 1 & 1 & 0 & 1 & 1 & 74.76 & 1.44 & 72.51 & 2.14 & 88.95 & 0.98 & 416 & 176 \\
        28 & 1 & 1 & 1 & 0 & 0 & 69.24 & 1.47 & 49.88 & 2.27 & 82.14 & 1.14 & 312 & 135 \\
        29 & 1 & 1 & 1 & 0 & 1 & 76.75 & 1.38 & 62.35 & 2.32 & 89.42 & 0.89 & 416 & 203 \\
        30 & 1 & 1 & 1 & 1 & 0 & 78.83 & 1.36 & 69.91 & 2.36 & 88.57 & 0.91 & 416 & 181 \\
        31 & 1 & 1 & 1 & 1 & 1 & 79.46 & 1.28 & 75.33 & 2.12 & 89.80 & 0.91 & 520 & 233 \\
        \hline
        \multicolumn{14}{@{}l}{\footnotesize{Filters:  V.--Variance  /  E.--Entropy  /  C.--Canny  /  G.--Gaussian / O.--Original}}\\
        \multicolumn{14}{@{}l}{\footnotesize{$\mu_{0}$--Avg. Success (Original Images)\qquad$\sigma_{0}$--Success Std. Deviation (Original Images)}}\\
        \multicolumn{14}{@{}l}{\footnotesize{$\mu_{PCA}$--Average Success (after PCA)\qquad$\sigma_{PCA}$--Success Standard Deviation (after PCA)}}\\
        \multicolumn{14}{@{}l}{\footnotesize{$\mu_{GA}$--Average Success (after G.O)\qquad$\sigma_{GA}$--Success Standard Deviation (after G.O.)}}\\
        \multicolumn{14}{@{}l}{\footnotesize{$NF_{0}$--Initial Number of Features\qquad$NF_{GA}$--Number of Features (after G.O.)}}
      \end{tabular}
\end{table}
\egroup

\begin{table*}[!ht]
  \centering
  \caption{Average confusion matrix for classification results on Case 23 for KCIMR -- CENPES Rock Dataset (in absolute values, not percentage). Rows represent the real classes of the images used while columns represent the prediction result of the classifier. High values in the diagonal of this matrix represent true positives, while values outside the diagonal represent misclassifications.}
  \label{tab:miscKCIMR}
      \begin{tabular}{| p{0.06\textwidth} | p{0.07\textwidth} p{0.07\textwidth} p{0.07\textwidth} p{0.07\textwidth} p{0.07\textwidth} p{0.07\textwidth} p{0.07\textwidth} p{0.07\textwidth} p{0.07\textwidth} |}
        \hline
    	\, & DPL & IBS & SD & IL & EYC & BBS & GBS & OLI & GNT \\
    	\hline 
        DPL & 107.53 & 0.10 & 0.72 & 1.49 & 2.11 & 0.00 & 0.05 & 0.00 & 0.04 \\
        IBS & \, & 99.08 & 4.20 & 6.34 & 0.63 & 1.95 & 7.53 & 5.75 & 2.56 \\
        SD & \, & \, & 88.11 & 1.60 & 19.19 & 0.01 & 0.46 & 7.07 & 0.26 \\
        IL & \, & \, & \, & 100.86 & 1.86 & 0.00 & 0.85 & 2.61 & 2.40 \\
        EYC & \, & \, & \, & \, & 100.29 & 0.00 & 0.11 & 4.63 & 0.01 \\
        BBS & \, & \, & \, & \, & \, & 107.17 & 2.45 & 1.87 & 0.70 \\
        GBS & \, & \, & \, & \, & \, & \, & 103.80 & 0.00 & 1.79 \\
        OLI & \, & \, & \, & \, & \, & \, & \, & 105.98 & 7.88 \\
        GNT & \, & \, & \, & \, & \, & \, & \, & \, & 105.94 \\
        \hline
    \end{tabular}
\end{table*}

\end{document}